\documentclass[letterpaper]{article} % DO NOT CHANGE THIS
\usepackage{aaai25}  % DO NOT CHANGE THIS
\usepackage{times}  % DO NOT CHANGE THIS
\usepackage{helvet}  % DO NOT CHANGE THIS
\usepackage{courier}  % DO NOT CHANGE THIS
\usepackage[hyphens]{url}  % DO NOT CHANGE THIS
\usepackage{graphicx} % DO NOT CHANGE THIS
\usepackage{natbib}  % DO NOT CHANGE THIS AND DO NOT ADD ANY OPTIONS TO IT
\usepackage{caption} % DO NOT CHANGE THIS AND DO NOT ADD ANY OPTIONS TO IT
\usepackage{algorithm}
\usepackage{algorithmic}
\usepackage{amsmath}
\usepackage{amsfonts}
\usepackage{subfigure}
\usepackage{booktabs}
\usepackage{xcolor}
\usepackage{newfloat}
\usepackage{listings}
\DeclareCaptionStyle{ruled}{labelfont=normalfont,labelsep=colon,strut=off} % DO NOT CHANGE THIS
\floatstyle{ruled}
\newfloat{listing}{tb}{lst}{}
\floatname{listing}{Listing}
\title{Spatial-Temporal Knowledge Distillation for Takeaway Recommendation}
\author{
    Shuyuan Zhao\textsuperscript{\rm 1,}\textsuperscript{\rm 2}\equalcontrib,
    Wei Chen\textsuperscript{\rm 1,}\textsuperscript{\rm 2}\equalcontrib, 
    Boyan Shi\textsuperscript{\rm 1,}\textsuperscript{\rm 2}, 
    Liyong Zhou\textsuperscript{\rm 1,}\textsuperscript{\rm 2}, 
    Shuohao Lin\textsuperscript{\rm 1,}\textsuperscript{\rm 2}, 
    Huaiyu Wan\textsuperscript{\rm 1,}\textsuperscript{\rm 2} \thanks{Corresponding authors.}\\
}
\affiliations{
    \textsuperscript{\rm 1}School of Computer Science \& Technology, Beijing Jiaotong University, Beijing, China\\
    \textsuperscript{\rm 2}Beijing Key Laboratory of Traffic Data Mining and Embodied Intelligence, Beijing, China\\
    \{sy\_zhao, w\_chen, by\_shi, zhouly, shhlin, hywan\}@bjtu.edu.cn
}

\usepackage{bibentry}
\begin{document}

\maketitle

\begin{abstract}
The takeaway recommendation system aims to recommend users' future takeaway purchases based on their historical purchase behaviors, thereby improving user satisfaction and boosting merchant sales.
Existing methods focus on incorporating auxiliary information or leveraging knowledge graphs to alleviate the sparsity issue of user purchase sequences.
However, two main challenges limit the performance of these approaches: (1) capturing dynamic user preferences on complex geospatial information and (2) efficiently integrating spatial-temporal knowledge from both graphs and sequence data with low computational costs.
In this paper, we propose a novel \textbf{s}patial-\textbf{t}emporal \textbf{k}nowledge \textbf{d}istillation model for takeaway \textbf{rec}ommendation (STKDRec) based on the two-stage training process.
Specifically, during the first pre-training stage, a spatial-temporal knowledge graph (STKG) encoder is trained to extract high-order spatial-temporal dependencies and collaborative associations from the STKG.
During the second spatial-temporal knowledge distillation (STKD) stage, a spatial-temporal Transformer (ST-Transformer) is employed to comprehensively model dynamic user preferences on various types of fine-grained geospatial information from a sequential perspective. 
Furthermore, the STKD strategy is introduced to transfer graph-based spatial-temporal knowledge to the ST-Transformer, facilitating the adaptive fusion of rich knowledge derived from both the STKG and sequence data while reducing computational overhead.
Extensive experiments on three real-world datasets show that STKDRec significantly outperforms 
the state-of-the-art baselines. 
% Our code is available at:  \text{https://github.com/Zhaoshuyuan0246/STKDRec}.
\end{abstract}

\begin{links}
    \link{Code}{https://github.com/Zhaoshuyuan0246/STKDRec}
    % \link{Datasets}{https://tianchi.aliyun.com/dataset/131047}
\end{links}

\section{Introduction}
Takeaway platforms, such as Yelp, Meituan, and Ele.me,
%\footnote{https://www.yelp.com/}
%\footnote{https://www.meituan.com/}
%\footnote{https://www.ele.me/}
provide convenient online ordering and offline delivery services, playing an increasingly important role in people's daily lives \cite{meituan}. 
As the core service of these platforms, takeaway recommendation aim to accurately recommend takeaways that align with user preferences based on their historical purchase behaviors.
Such recommendation services enhance user satisfaction while increasing visibility and sales opportunities for merchants. 
In recent years, takeaway recommendations have gained significant attention from the research community and industry \cite{StEN, BASM, shi2024self}. 

\begin{figure}[t]
  \centering
    \includegraphics[width=1 \linewidth]{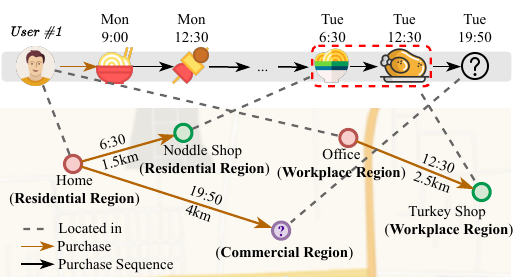}
  \caption{An illustrative example highlighting the importance of capturing dynamic user preferences on complex geospatial information.}
\end{figure}

Takeaway recommendation is essentially a sequential modeling task that aims to predict the user's future takeaway purchases based on their historical purchase records.
Existing methods model these purchase sequences by utilizing deep neural networks, such as recurrent neural networks (RNNs) and self-attention mechanisms. 
However, the sparsity issue arises since users often purchase only a few takeaways, limiting the performance of recommendations \cite{kang2023distillation}.
Some studies attempt to incorporate various types of auxiliary information, such as brand, category, location, and area of interest (AOI), into purchase sequences. 
Some methods, such as KGDPL \cite{KGDPL}, also utilize graph neural networks (GNNs) and Knowledge Graphs (KGs) \cite{, wu2023towards, chen2024local} to explore higher-order user-takeaway relationships or rich semantics of takeaways, alleviating the data sparsity problem.
Although these methods all achieve promising performance, two significant challenges still need to be addressed:

\textbf{(1) Failing to effectively capture dynamic user preferences on complex geospatial information.} 
In takeaway recommendation scenarios, user preferences change dynamically over time and their current location. 
For example, users tend to purchase fast food at noon when at the workplace, while they prefer main meals in the evening when at home. 
However, complex geospatial information, including the distance between the user and the delivery merchant, the functional region of the merchant, and so on, as important factors affecting user preferences, has not been adequately explored.
As illustrated in Figure 1, \textit{User \#1} is primarily located in a workplace region during the day, frequently purchasing food that matches their preferences from nearby shops.
However, when the user returns to the residential region in the evening, existing methods prioritize recommending nearby food candidates. 
Satisfying user preferences for foods from a more distant commercial region is neglected due to insufficient consideration of complex geospatial information. 
Therefore, capturing dynamic user preferences on complex geospatial information poses a challenge.

\textbf{(2) How to efficiently integrate spatial-temporal knowledge from both graphs and sequence data with low computational costs.} 
The user's purchase history is sequential data, while KGs are non-Euclidean structure data.
Effectively integrating the advantages of these two types of heterogeneous data can alleviate the challenge of data sparsity and improve the accuracy of recommendations.
However, due to the typically large scale of KGs, encoding them with GNNs significantly increases computational overhead.
Additionally, simple knowledge fusion methods, such as addition or concatenation, are not conducive to integrating these heterogeneous data for subsequent recommendations. 
Therefore, finding an effective method to fuse spatial-temporal knowledge from both graphs and sequence data while reducing computational costs is crucial.

To address these challenges, we propose a novel \textbf{S}patial-\textbf{T}emporal \textbf{K}nowledge \textbf{D}istillation model for takeaway \textbf{rec}ommendation, termed \textbf{STKDRec}. 
The model distills the offline teacher model's knowledge of the graph structure to better enhance the student model's ability to model users' historical purchase sequences while improving computational efficiency.
STKDRec consists of two stages: pre-training and spatial-temporal knowledge distillation (STKD). 
During the pre-training stage, STKDRec constructs a spatial-temporal knowledge graph (STKG) and extracts high-order spatial-temporal dependencies and collaborative associations between users and takeaways from STKG through training an STKG encoder. 
During the STKD stage, a spatial-temporal Transformer (ST-Transformer) is presented to capture dynamic user preferences on various types of fine-grained geospatial information from a sequential perspective. 
Through the STKD strategy, graph-based spatial-temporal knowledge from the STKG encoder is effectively transferred to the ST-Transformer, facilitating heterogeneous knowledge fusion with low computational costs.

The contributions of this work are as follows:
\begin{itemize}
\item We propose a novel spatial-temporal knowledge distillation model for takeaway recommendations, utilizing knowledge distillation to fuse spatial-temporal knowledge from both STKG and sequence data, thereby addressing the sparsity issue in sequence data and effectively reducing computational overhead.

\item We integrate various types of geospatial information into user sequence data and propose an ST-Transformer to capture dynamic user preferences on complex geospatial information from a sequential perspective.

\item Extensive experiments on three real-world datasets 
show that STKDRec achieves superior recommendation performance over the state-of-the-art baselines.
\end{itemize}

\begin{figure*}[htbp]
    \centering
    \includegraphics[width=0.95 \linewidth]{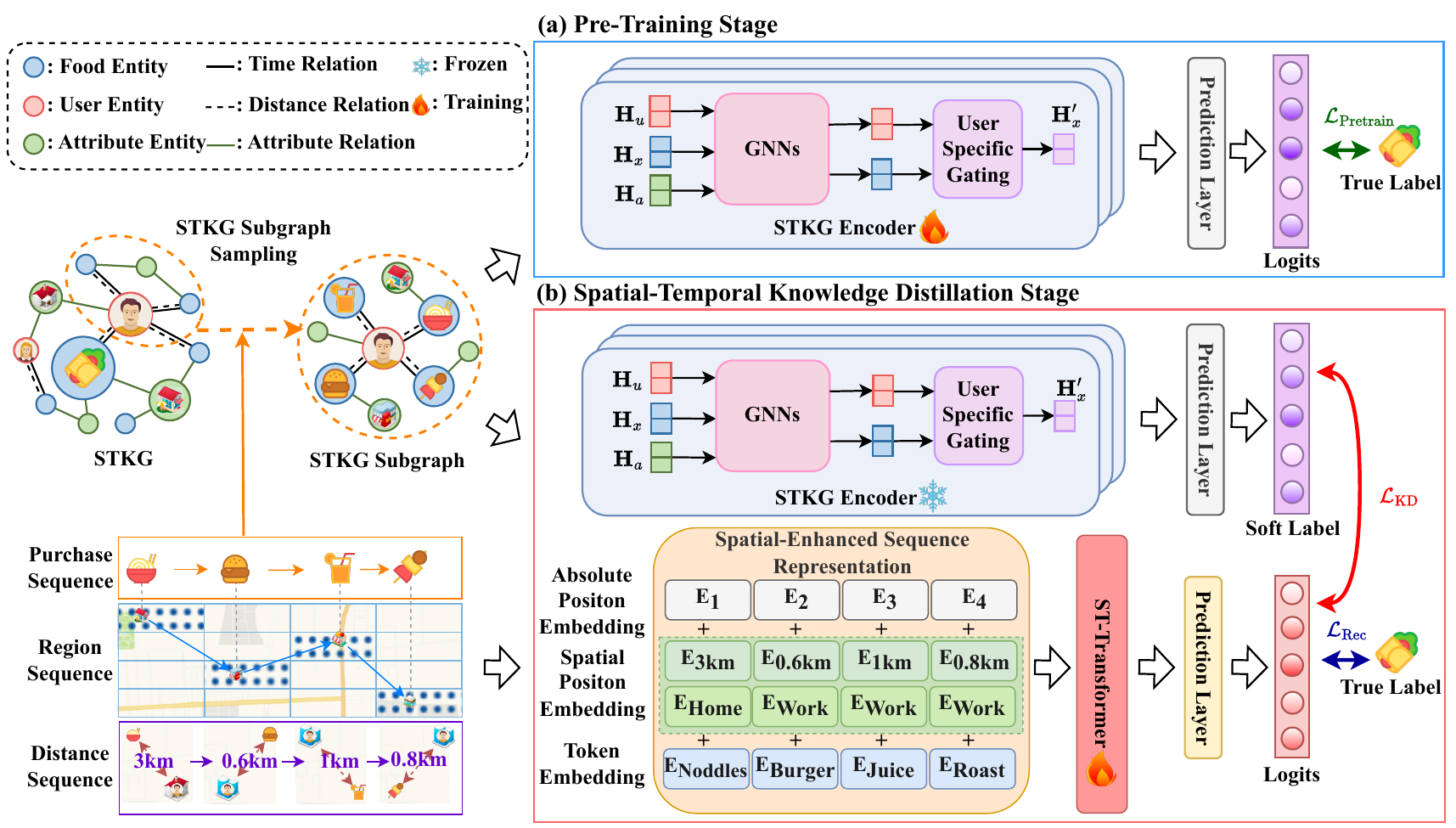}
    \caption{The overall architecture of STKDRec, consisting of two stages: the pre-training stage and the STKD stage.}
    \label{fig3}
\end{figure*}

\section{Related Work}
\subsection{Takeaway Recommendation}
Takeaway recommendation methods recommend personalized takeaways to users based on their historical purchase sequences.
Early methods used Markov Chains (MC) \cite{Markov_model} to model user sequences. With the development of deep learning, methods such as GNNs, RNNs, and self-attention mechanisms \cite{GRU4Rec, GCL4SR, chen2022building, BSARec} have been used to model users' time-varying interests.
Some methods attempt to capture users' spatial-temporal interests by considering spatial-temporal information.
StEN \cite{StEN} models the spatial-temporal information of users and foods for click-through rate prediction in location-based service. 
BASM \cite{BASM} integrates spatial-temporal features to capture user preferences at different times and locations. 
However, these methods primarily focus on spatial region information (e.g., geohash or AOI) without accounting for fine-grained spatial information of different types, such as spatial distance between the users and the takeaways. 
They fail to comprehensively and accurately capture dynamic user preferences in takeaway recommendations.
Therefore, we propose a spatial-enhanced sequence representation encompassing various types of geospatial information and utilize an ST-Transformer to learn dynamic user preferences from a
sequential perspective.

\subsection{Knowledge Distillation}
Knowledge distillation \cite{hinton2015distilling}  achieves lightweight and performance enhancement by transferring knowledge from a teacher model to a student model.
Existing methods \cite{kang2021topology, xia2022device, zhu2021combining} can be categorized into distillation between models of the same type and distillation between models of different types. 
Distillation between models of the same type aims to reduce parameters, enabling the student model to approach the teacher model's performance with fewer parameters. 
For example, TinyBERT \cite{devlin2018bert} uses the BERT model for pre-training and fine-tuning, distilling it into the smaller language model. 
Distillation can also be performed between different types of models.
When there are structural or mechanistic differences between the models, the teacher model can provide a unique knowledge background and global understanding that are difficult for the student model to obtain independently.
For instance, Graphless Neural Network \cite{GLNN} translates the knowledge from the graph structure of GNNs into the MLP format.
Motivated by these approaches, we leverage knowledge distillation to transfer the graph-based spatial-temporal knowledge from the STKG encoder to the ST-Transformer, thereby achieving the purpose of heterogeneous knowledge fusion and reducing computational costs.

\section{Preliminaries}
\subsubsection{Problem Formulation}
The goal of takeaway recommendation is to predict the users' next takeaway purchase based on their historical purchase sequences.
Given a set of users $\mathcal{U}$ and a set of takeaways $\mathcal{V}$, we can sort the purchased takeaways of each user $u \in \mathcal{U}$ chronologically in a sequence as $x = (v_1, v_2, \dots, v_{|x|})$, where $v_i \in \mathcal{V}$ denotes the $i$-th purchased takeaway in the sequence.
The task is to recommend a Top-$k$ list of takeaways as potential candidates for the user's next purchase.
Formally, we predict $P(v_{|x|+1} \mid x).$

\subsubsection{Spatial-Temporal Knowledge Graph}
The STKG is represented as $\mathcal{G} = \mathcal{(E, R, T)}$, which consists of a set of triples composed of entity-relation-entity.
Here, $\mathcal{E}$, $\mathcal{R}$, and $\mathcal{T}$ denote the entity set, the relation set, and the triple set, respectively.
In the STKG, entities include users, takeaways, and their associated attributes, while relations encompass \textit{time relation}, \textit{distance relation}, and \textit{attribute relation}.
These entities and relations form four different types of triples: time triples of the user purchasing the takeaway $(u, \textit{time}, v)$, distance triples of the user purchasing the takeaway $(u, \textit{distance}, v)$, user-attribute triples $(u, \textit{attribute}, a)$, and takeaway-attribute triples $(v, \textit{attribute}, a)$.

\section{Our Approach}
The overall framework of STKDRec is shown in Figure 2, which consists of two stages: the pre-training stage and the STKD stage. 
During the pre-training stage, an STKG encoder is trained to model the high-order spatial-temporal dependencies and collaborative associations between users and takeaways from the perspective of the graph structure.
During the STKD stage, an ST-Transformer is employed to model the dynamic user preferences on complex spatial information from a sequential perspective. The STKD strategy facilitates the fusion of spatial-temporal knowledge from both STKG and sequence data.

\subsection{Spatial-Temporal Knowledge Graph Encoder}
The STKG encoder, as the teacher model, is pre-trained to capture high-order spatial-temporal knowledge from the STKG.
% enabling it to model spatial-temporal dependencies and collaborative associations between users and takeaways from a graph structure perspective.
Thus, we sample an STKG subgraph from STKG based on the user purchase sequence. The STKG encoder is used to aggregate spatial-temporal knowledge from the subgraph while integrating personalized user features.

\subsubsection{STKG Subgraph Sampling} 
% We sample the corresponding STKG subgraph based on the user purchase sequence. 
In this work, we use the efficient neighborhood sampling method \cite{STKGsample} to sample the subgraph from the STKG.
Specifically, given a user’s purchase sequence $x$  and a maximum sequence length $n$, the sequence is truncated by removing the earliest takeaways if $|x| > n$ or padded with 0 to get a fixed length sequence $x = (v_1, v_2, \dots, v_n)$. 
Each node $v_i$ in $x$ is treated as a center node, and a fixed number $s$ of neighbor nodes are randomly sampled from its adjacent nodes in the STKG.
This process retains the relationships between $v_i$ and sampled neighbors.
The same process is recursively applied for each neighbor node to sample additional adjacent nodes and relationships, continuing to a specified depth $m$.
All center nodes, their sampled neighbors, and the retained relations are then integrated into an STKG subgraph, denoted as $\mathcal{G}_x$.

\subsubsection{Spatial-Temporal Knowledge Aggregation}
To model the higher-order spatial-temporal dependencies and collaborative associations between users and takeaways with $\mathcal{G}_x$, we utilize the GNNs to effectively encode the $\mathcal{G}_x$.
In the $l$-th layer of the GNNs, the message passing and aggregation processes are defined as follows:
\begin{equation}
\begin{split}
m_{v_i}^{l} = \text{Aggregate}^{l}\left(\{\{{h}_{v_j}^{(l-1)}, h_r\}: \exists\left(v_{i}, r, v_{j}\right) \in \mathcal{G}_{x}\}\right),
\end{split}
\end{equation}
\begin{equation}
h_{v_i}^{l} = \text{Combine}^{l}\left(m_{v_i}^{l}, h_{v_i}^{(l-1)}\right),
\end{equation}
where $h_{v_{i}}^{(l-1)}$ and $h_{v_j}^{(l-1)}$ represent the embeddings of entity  $v_i$ and its neighboring entity $v_j$ at the layer $(l-1)$, respectively. 
$h_r$ represents the embedding of relationship $r \in \mathcal{R}$ between $v_i$ and $v_j$.
$m_{v_i}^{l}$ denotes the representation of aggregated neighborhood message for $v_i$ at layer $l$. $\text{Aggregate(·)}$ is a function that aggregates the neighborhood information of the central node $v_i$, while $\text{Combine(·)}$ merges this information to update the entity embeddings. 
After propagation information through multiple GNN layers on $\mathcal{G}_x$, we obtain the final embeddings of all entities in $x$, denoted as $\text{H}_x \in \mathbb{R}^{n \times d}$, and the final embedding for user $u$, denoted as $\text{H}_u \in \mathbb{R}^{1 \times d}$, where $d$ is the latent dimension.

To model users' specific spatial-temporal preferences, we introduce a user specific gating mechanism to incorporate personalized user features into $\text{H}_x$. The user specific representation $\text{H}_x'$ is defined as:
\begin{equation}
\text{H}_x' = \text{H}_x \otimes \sigma( \text{H}_x \text{W}_1 + \text{W}_2 \text{H}_u^\top ),    
\end{equation} 
where $\text{W}_1 \in \mathbb{R}^{d \times 1}$ and $\text{W}_2 \in  \mathbb{R}^{n \times d}$ represent learnable parameters, $\sigma\text{(·)}$ denotes the sigmoid activation function, $\otimes$ indicates element-wise multiplication.

\subsubsection{Soft Labels}
% \subsubsection{Prediction Layer}
% The STKG encoder as a teacher model in STKDRec, captures high-order spatial-temporal and collaboration associations between users and takeaways from the perspective of graph structure.
To facilitate the subsequent STKD stage, we use the prediction distribution generated by the STKG encoder as soft labels. These labels guide the student model in learning the high-order spatial-temporal knowledge from the STKG. Specifically, after deriving $\text{H}_x'$, an attention mechanism is applied to obtain the representation of $x$. The soft labels are formally defined as:
% To facilitate subsequent STKD, we generate the distribution of prediction as soft labels for the student model. Specifically, given the purchase sequence $x$ of the user $u$ with length $n$, the purchase probability between the user and takeaways at $(n+1)$ step based on $\text{H}'_{x, n}$ can be defined as:
\begin{equation}
    \text{Y}_{x}' = \text{Softmax} (\text{AttNet}(\text{H}'_{x})  \text{E}_\mathcal{V}^\top),
\end{equation}
where $\text{Y}_x' \in \mathbb{R}^{1 \times |\mathcal{V}|}$, the $i$-th element of $\text{Y}_x'$ represents the purchase probability of the $i$-th takeaway, $\text{AttNet(·)}$ denotes the attention network, and $\text{E}_\mathcal{V} \in \mathbb{R}^{|\mathcal{V}| \times d}$ represents the embedding of all takeaways.

\subsection{Spatial-Temporal Transformer}
The ST-Transformer, as the student model, is designed to model dynamic user preferences on complex geospatial information from a sequential perspective. 
In real-world scenarios, complex geospatial information (e.g., spatial regions and spatial distances) significantly influences user preferences. Spatial region reflects users' general preferences, while spatial distance reveals users' specific preferences across regions. 
To model these various types of geospatial information, we introduce a spatial-enhanced sequence representation that integrates these diverse geospatial factors, enabling the ST-Transformer to learn dynamic user preferences that vary across these factors.
% By jointly considering these various types of geospatial information, the model effectively understands users' complex needs. 
% Thus, we introduce a spatial-enhanced sequence representation that integrates these diverse geospatial factors, enabling the ST-Transformer to learn dynamic user preferences that vary across these factors.
% Geospatial information, such as spatial region reflects users' general preferences, while spatial distance information reveals specific preferences across regions.
% By jointly considering spatial regions and distances, the model effectively understands users’ complex needs.
% Thus, we introduce a spatial-enhanced sequence representation that integrates spatial regions and distances, and utilize the ST-Transformer to learn dynamic preferences that vary across these factors.

\subsubsection{Spatial-Enhanced Sequence Representation}
To utilize the sequential order of tokens in the sequence, previous works \cite{Bert4Rec, gao2023learning} add an absolute position embedding $\text{E}_\text{P}$ to enhance the sequence.
Inspired by these, we propose a novel spatial position embedding to enhance user purchase sequences, enabling the ST-Transformer to focus not only on the users' evolving interests over time but also on how these interests shift across various geospatial information. 
In our approach, we integrate spatial regions and spatial distances to construct the spatial position embeddings. 
Specifically, the spatial region set $\mathcal{C}$ encompasses predefined geohash6 attributes \cite{BASM} of all takeaways, and the spatial distance set $\mathcal{F}$ encompasses distances between the regions of users and the regions of the takeaways.
Given the sequence $x$ of the user $u$ with length $n$, the embeddings matrix of all takeaways in $x$ is denoted as $\text{E}_x \in \mathbb{R}^{n \times d}$. 
Similarly, the spatial regions embedding matrix $\text{E}_{x_c}  \in \mathbb{R}^{n \times d}$ and the spatial distance embedding matrix $\text{E}_{x_f}  \in \mathbb{R}^{n \times d}$ are defined based on the spatial region sequence $x_c = (c_1, c_2, \dots, c_n)$ and the spatial distance sequence $x_f = (f_1, f_2, \dots, f_n)$, where $c_i \in \mathcal{C}$ and $f_i \in \mathcal{F}$.
By combining $\text{E}_{x_c}$ with $\text{E}_{x_f}$, we obtain a learnable spatial position embedding $\text{E}_\text{SP}$:
\begin{equation}
    \text{E}_\text{SP} = \text{W}_\text{SP} (\text{E}_{x_c} + \text{E}_{x_f}),
\end{equation}
where $\text{W}_\text{SP}$ represents a learnable parameter. 
Finally, we sum the three embeddings $\text{E}_x,\text{E}_\text{SP},\text{E}_\text{P}$ to produce the spatial-enhanced sequence representation $\hat{\text{E}}_x \in \mathbb{R}^{n \times d}$,
\begin{equation}
    \hat{\text{E}}_x =\text{E}_x + \text{E}_\text{SP} + \text{E}_\text{P}.
\end{equation}

\subsubsection{Spatial-Temporal Context Attention}
To extract dynamic user preferences that vary across regions and distances from spatial-enhanced sequence representation, we present the spatial-temporal context attention mechanism. 
Specifically, this mechanism comprises $L$ layers of mask self-attention layers stacked together, transforming the input embedding $\hat{\text{E}}_x$ into the spatial-temporal context representation $\hat{\text{H}}_x \in \mathbb{R}^{n \times d}$.
In each layer, we employ three independent linear transformation matrices $\text{W}_i^q, \text{W}_i^k, \text{W}_i^v \in \mathbb{R}^{d \times d'}$ to transform the input embedding $\hat{\text{E}}_x$ into queries, keys, and values for the $i$-th scaled dot-product attention head, where $d' = \frac{d}{K}$, and $i = 1, 2, \ldots, K$.
The function is defined as follows:
\begin{equation}
    \hat{\text{h}}_i = \text{Softmax}\left(\frac{(\hat{\text{E}}_x \text{W}_i^q)(\hat{\text{E}}_x \text{W}_i^k)^\top}{\sqrt{d'}}\right)(\hat{\text{E}}_x \text{W}_i^v),
\end{equation}
where $\hat{\text{h}}_i \in \mathbb{R}^{n \times d'}$ denotes the output representation of the corresponding attention head. We concatenate the outputs from all attention heads to obtain the final spatial-temporal context representation $\hat{\text{H}}_x$,
\begin{equation}
    \hat{\text{H}}_x = \text{FFN}(\text{Concatenate}(\hat{\text{h}}_1, \hat{\text{h}}_2, \dots \hat{\text{h}}_K)),
\end{equation}
where $\text{FFN(·)}$ denotes the feed-forward network.

\subsubsection{\textbf{Prediction Layer}}
To achieve the takeaway recommendation task, the final spatial-temporal context representation $\hat{\text{H}}_{x}$ 
is multiplied by the embeddings $\text{E}_\mathcal{V}$ of all takeaways to predict the probability of takeaway appearing at $(n+1)$ step:
\begin{equation}
    \hat{\text{Y}}_x = \text{Softmax}(\hat{\text{H}}_{x} \text{E}_\mathcal{V}^\top),
\end{equation} 
where the $j$-th element of $\hat{\text{Y}}_x$ denotes the purchase probability of the $j$-th takeaway.
% where $\hat{\text{Y}}_x \in \mathbb{R}^{1 \times |\mathcal{V}|}$, the $j$-th element of $\hat{\text{Y}}_x$ denotes the purchase probability of the $j$-th takeaway.

\subsection{Spatial-Temporal Knowledge Distillation}
Although the ST-Transformer captures dynamic user preferences from a sequential perspective, it fails to capture the high-order spatial-temporal dependencies and collaborative associations between users and takeaways from the STKG. Thus, we propose the STKD strategy, which facilitates the transfer of graph-based spatial-temporal knowledge from the STKG encoder to the more efficient and lightweight ST-Transformer.
This approach enables the heterogeneous fusion of spatial-temporal knowledge while significantly reducing the computational costs.

During the pre-training stage, the teacher model STKG encoder supervises the learning process using ground truth labels $\text{Y}_x \in \mathbb{R}^{1 \times |\mathcal{V}|}$, generating soft labels $\text{Y}_x'$ following pre-training. The pre-training loss is defined as:
\begin{equation}
   \mathcal{L}_{\text{Pretrain}} = \text{CrossEntropy}(\text{Y}_x', \text{Y}_x). 
\end{equation}

During the STKD stage, our goal is to distill valuable spatial-temporal knowledge from the STKG encoder, thereby enhancing the ST-Transformer's ability to capture user preferences from different perspectives and promoting more efficient, streamlined learning.
To achieve this, we train the student model, ST-Transformer, to emulate the soft labels of the teacher model, effectively transferring knowledge from the teacher to the student.
The distillation loss is defined as:
\begin{equation}
    \mathcal{L}_{\text{KD}}({\text{Y}_x'}, \hat{\text{Y}}_x) = \text{KL}\left(\text{Y}_x'/\tau \, \Big|\Big| \, \hat{\text{Y}}_x/\tau\right),
\end{equation}
where KL denotes the Kullback-Leibler divergence, and $\tau$ is the temperature coefficient.
Furthermore, to ensure that the student model also learns the true labels, we optimize a supervised loss:
\begin{equation}
\mathcal{L}_{\text{Rec}} = \text{CrossEntropy}(\hat{\text{Y}}_x, \text{Y}_x).
\end{equation}
Ultimately, the student model is trained by jointly optimizing the distillation loss and the supervised loss:
\begin{equation}
    \mathcal{L} = \alpha \mathcal{L}_{\text{KD}} + (1-\alpha)\mathcal{L}_{\text{Rec}}. 
\end{equation}
where $\alpha \in [0,1]$ is the coefficient to balance the distillation loss and the supervised loss.

\section{Experiments}
\subsection{Experimental Setting}
\subsubsection{Datasets}
We select three publicly available city takeaway recommendation datasets for evaluation: Wuhan, Sanya, and Taiyuan, which are provided by the well-known takeaway platform Ele.me.
% \footnote{The dataset is available at \url{https://tianchi.aliyun.com/dataset/131047}.}.
Each dataset contains users' purchase sequences and associated attributes.
The attributes mainly include geospatial features, such as the geohash6 and AOI, where users and takeaways are located, temporal features, such as the timestamps and weekdays when users purchase takeaways, and additional user and takeaway attributes.
To ensure the quality of data, we apply the following processing steps. 
Initially, we remove the users and takeaways with empty geohash6 attribute values to clean the dataset.
Secondly, we retain features from the cleaned dataset and calculate the spherical distance as a spatial distance feature based on the geohash6 values of users and takeaways.
Finally, for each user purchase sequence, the last purchased takeaway is designated as test data, the second-to-last as validation data, and the rest as training data. A statistical summary of the processed datasets is shown in Table 1.

\begin{table}[ht]
\setlength\tabcolsep{2.5pt}%调列距
\centering
\scriptsize
\begin{tabular}{ccccccc}
\toprule
City & \# Samples & \# Users & \# Takeaways & \# Entities & \# Relations & Avg. length \\
\midrule
Wuhan & 2,125,761 & 31,047 & 383,953 & 447,033 & 4,906 & 41.27\\
Sanya & 3,479,272 & 19,842 & 168,332 & 234,307 & 3,795 & 42.02\\
Taiyuan & 4,442,984 & 32,412 & 306,344 & 396,153 & 3,669 & 39.97\\
\bottomrule
\end{tabular}
\caption{Statistics of the processed datasets.}
\label{tab:datasets}
\end{table}

\subsubsection{Baselines}
% 20240810 17:40修订
To evaluate the effectiveness of our model, we compare it with the following nine representative baselines: Caser \cite{Caser}, GRU4Rec \cite{GRU4Rec}, SASRec \cite{SASRec}, BERT4Rec \cite{Bert4Rec}, DuoRec \cite{DuoRec}, FEARec \cite{FEARec},  GCL4SR \cite{GCL4SR}, MAERec \cite{MAERec}, and BSARec \cite{BSARec}.

\begin{table*}[ht]
\setlength\tabcolsep{2pt}%调列距
\renewcommand{\arraystretch}{0.89}
    \centering	
    \scalebox{1}{
    \begin{tabular}{cccccccccccccc}
        \toprule
        \textbf{Dataset} & \textbf{Metric} & \textbf{Caser} & \textbf{GRU4Rec} & \textbf{BERT4Rec} & \textbf{MAERec} & \textbf{DuoRec} & \textbf{FEARec} & \textbf{SASRec} & \textbf{GCL4SR} & \textbf{BSARec} & \textbf{STKDRec} \\
        \midrule
        & HR@5 & 0.5797 & 0.7530 & 0.6299 & 0.7441 & \underline{0.7659} & 0.7652 & 0.7420 & 0.7625 & 0.7643 & \textbf{0.7909} \\
        & HR@10 & 0.6534 & 0.7854 & 0.6887 & \underline{0.7976} & 0.7913 & 0.7904 & 0.7796 & 0.7938 & 0.7958 & \textbf{0.8229} \\
        {Wuhan} & HR@20 & 0.7225 & 0.8120 & 0.7485 & 0.8170 & 0.8141 & 0.8133 & 0.8160 & 0.8242 & \underline{0.8252} & \textbf{0.8659} \\
        & NDCG@5 & 0.4690 & 0.6806 & 0.5631 & 0.6245 & 0.7210 & 0.7198 & 0.6789 & 0.7159 & \underline{0.7216} & \textbf{0.7463} \\
        & NDCG@10 & 0.4929 & 0.6943 & 0.5821 & 0.6485 & 0.7292 & 0.7279 & 0.6911 & 0.7261 & \underline{0.7318} & \textbf{0.7586} \\
        & NDCG@20 & 0.5104 & 0.7036 & 0.5972 & 0.6661 & 0.7350 & 0.7338 & 0.7003 & 0.7340 & \underline{0.7393} & \textbf{0.7694} \\
        \midrule
        & HR@5 & 0.8338 & 0.8580 & 0.7965 & 0.7983 & 0.8561 & 0.8578 & 0.8560 & 0.8566 & \underline{0.8623} & \textbf{0.8770} \\
        & HR@10 & 0.8679 & 0.8715 & 0.8294 & 0.8612 & 0.8747 & 0.8769 & 0.8806 & 0.8825 & \underline{0.8838} & \textbf{0.8940} \\
        {Sanya} & HR@20 & 0.8931 & 0.8894 & 0.8627 & 0.8862 & 0.8884 & 0.8908 & 0.8987 & 0.8992 & \underline{0.9005} & \textbf{0.9106} \\
        & NDCG@5 & 0.7326 & 0.7967 & 0.7410 & 0.6478 & 0.8072 & 0.8095 & 0.8004 & 0.8060 & \underline{0.8189} & \textbf{0.8437}\\
        & NDCG@10 & 0.7446 & 0.8044 & 0.7514 & 0.6750 & 0.8133 & 0.8157 & 0.8084 & 0.8145 & \underline{0.8259} & \textbf{0.8492}\\
        & NDCG@20 & 0.7512 & 0.8089 & 0.7594 & 0.6840 & 0.8168 & 0.8192 & 0.8130 & 0.8217 &\underline{0.8301} & \textbf{0.8534}\\
        \midrule
        & HR@5 & 0.7329 & 0.8259 & 0.7616 & 0.8112 & 0.8343 & 0.8345 & 0.8319 & \underline{0.8541} & 0.8419 & \textbf{0.8630} \\
        & HR@10 & 0.7916 & 0.8543 & 0.7985 & 0.8540 & 0.8564 & 0.8533 & 0.8583 & 0.8637 & \underline{0.8661} & \textbf{0.8789} \\
        {Taiyuan} & HR@20 & 0.8396 & 0.8769 & 0.8351 & 0.8781 & 0.8761 & 0.8738 & 0.8813 & 0.8869 & \underline{0.8881} & \textbf{0.8963} \\
        & NDCG@5 & 0.6175 & 0.7900 & 0.7115 & 0.6965 & 0.7923 & 0.7923 & 0.7774 & 0.8049 & \underline{0.8053} & \textbf{0.8340}\\
        & NDCG@10 & 0.6366 & 0.7982 & 0.7235 & 0.7138 & 0.7994 & 0.7984 & 0.7860 & 0.8099 & \underline{0.8132} & \textbf{0.8391}\\
        & NDCG@20 & 0.6488 & 0.8040 & 0.7327 & 0.7225 & 0.8044 & 0.8036 & 0.7918 & 0.8162 & \underline{0.8188} & \textbf{0.8435}\\
        \bottomrule
    \end{tabular}
    }
    \caption{Overall performance comparison. The best results are in boldface and the second-best results are underlined.}
    \label{tab:results}
\end{table*}

\subsubsection{Implementation Details}
All evaluation methods are implemented in PyTorch. The hyper-parameters for these methods are chosen according to the original papers, and the optimal settings are selected based on model performance on the validation data. For STKDRec, we conduct experiments with the following hyper-parameters. For the STKG encoder, the sampling depth $m$ is set to 2, and the number of neighbor nodes sampled $s$ is selected from \{[5,5], [10,10], [15,15], [20,20]\}. For the ST-Transformer module, we set the number of self-attention blocks and attention heads to 2 and the embedding dimension to 256. For STKD, the temperature $\tau$ is selected from \{1, 3, 5, 7, 9\}, and the $\alpha$ is fixed at 0.2. We use Adam as the optimizer, with the learning rate, $\beta_1$, and $\beta_2$ set to 0.001, 0.9, and 0.98, respectively. The batch size and maximum sequence length $n$ are both set to 128. An early stopping strategy is applied based on the performance of the validation data. The implementation is carried out in PyTorch on a single NVIDIA A40 GPU with 48GB of memory.

\subsubsection{Metrics}
To measure the accuracy of recommendations, we use the widely adopted Top-$k$ metrics HR@$k$ (Hit Rate) and NDCG@$k$ (Normalized Discounted Cumulative Gain), with $k$ set to 5, 10, and 20. HR@$k$ calculates the frequency with which the actual next takeaway appears within the top-$k$ recommendations, while NDCG@$k$ is a position-aware metric assigning higher weights to takeaways ranked higher. To evaluate STKDRec, we pair the actual takeaway from the test set with 100 randomly sampled negative takeaways that the user has not purchased and subsequently rank them.

\subsection{Experimental Results}
Table 2 presents the overall experimental results of STKDRec and baselines. We have the following vital observations: STKDRec achieves the best performance across all datasets compared to all baselines, highlighting the effectiveness of our STKDRec. The second-best model BSARec considers fine-grained user sequential patterns without accounting for spatial-temporal dependencies and collaborative associations between users and takeaways. These limitations result in BSARec performing worse than STKDRec.
% (2) Compared to BSARec, which rely solely on sequence data, STKDRec significantly outperformed them. This enhancement primarily results from these methods relying exclusively on user behavior sequences, which limits their ability to capture users' dynamic geospatial preferences. STKDRec effectively captures dynamic user preferences across varying geospatial information by integrating diverse geospatial information into user behavior sequences.

Graph-based methods GCL4Rec and MAERec focus on user-item graphs to capture the collaborative association between users and takeaways but lack the integration of temporal, spatial, and additional auxiliary information, limiting their ability to capture users' preferences over time and geospatial information. In takeaway recommendation scenarios, these factors significantly influence users' purchase behavior. This is why GCL4SR and MAERec perform worse than STKDRec. 
% This demonstrates the necessity of incorporating diverse temporal, spatial, and attribute information in STKDRec.
Feature-enhanced methods DuoRec and FeaRec rely on static features or limited external knowledge, making it difficult to effectively capture the dynamic changes in user preferences. Our STKDRec captures dynamic user preferences on complex geospatial information and integrates the spatial-temporal knowledge within the STKG.
% (3) Compared to graph-based methods such as GCL4SR and MAERec, STKDRec provides superior performance. These methods construct a graph lacking spatial and temporal information integration, which limits their effectiveness. The STKG introduced by STKDRec includes diverse temporal, spatial, and attribute relations, allowing the model to effectively model high-order spatial-temporal and collaborative associations between users and takeaways.
% (4) Compared to feature-enhanced methods such as DuoRec and FeaRec, STKDRec also performs exceptionally well. These methods rely on static features or limited external knowledge, failing to capture the dynamic user preferences over spatial-temporal information. STKDRec overcomes these limitations by introducing the STKG and by incorporating diverse geospatial information to enhance user behavior.

\begin{table}[ht]
\setlength\tabcolsep{4pt}%调列距
\scriptsize
\renewcommand{\arraystretch}{0.9}
    \centering
        \scalebox{1}{
    \begin{tabular}{cccccccccc}
        \toprule
        {\textbf{Method}} & \multicolumn{2}{c}{\textbf{Wuhan}} & & \multicolumn{2}{c}{\textbf{Sanya}} & & \multicolumn{2}{c}{\textbf{Taiyuan}} \\
        \cmidrule(lr){2-3} \cmidrule(lr){5-6} \cmidrule(lr){8-9}
        & \textbf{H@10} & \textbf{N@10} & & \textbf{H@10} & \textbf{N@10} & & \textbf{H@10} & \textbf{N@10}\\
        \midrule
        \textit{-w/o SP+KD} & 0.7796 & 0.6911 & &  0.8806 & 0.8084 & & 0.8583 & 0.7860\\
        \textit{-w/o KD} & 0.8161 & 0.7509 & & 0.8838 & 0.8294 & & 0.8684 & 0.8277\\
        \textit{-w/o SP} & 0.8097 & 0.7456 & & 0.8844 & 0.8348 & & 0.8669 & 0.8266\\
        \textit{-w/o C} & 0.8181 & 0.7518 & & 0.8875 & 0.8379 & & 0.8685 & 0.8267 \\
        \textit{-w/o F} & 0.8178 & 0.7535 & & 0.8877 & 0.8384 & & 0.8688 & 0.8277 \\
        % w/o LM & 0.8737 & 0.8226 & & 0.8885 & 0.8245 \\
        \midrule
        STKDRec & \textbf{0.8229} & \textbf{0.7586} & & \textbf{0.8940} & \textbf{0.8492} & & \textbf{0.8789} & \textbf{0.8391} \\
        \bottomrule
    \end{tabular}
    }
    \caption{The results of ablation studies.}
    \label{tab:results2}
\end{table}

\begin{figure}[ht]
\centering
\subfigure{
    \begin{minipage}[t]{0.47\linewidth}
        \centering
        \includegraphics[width=1.05\linewidth]{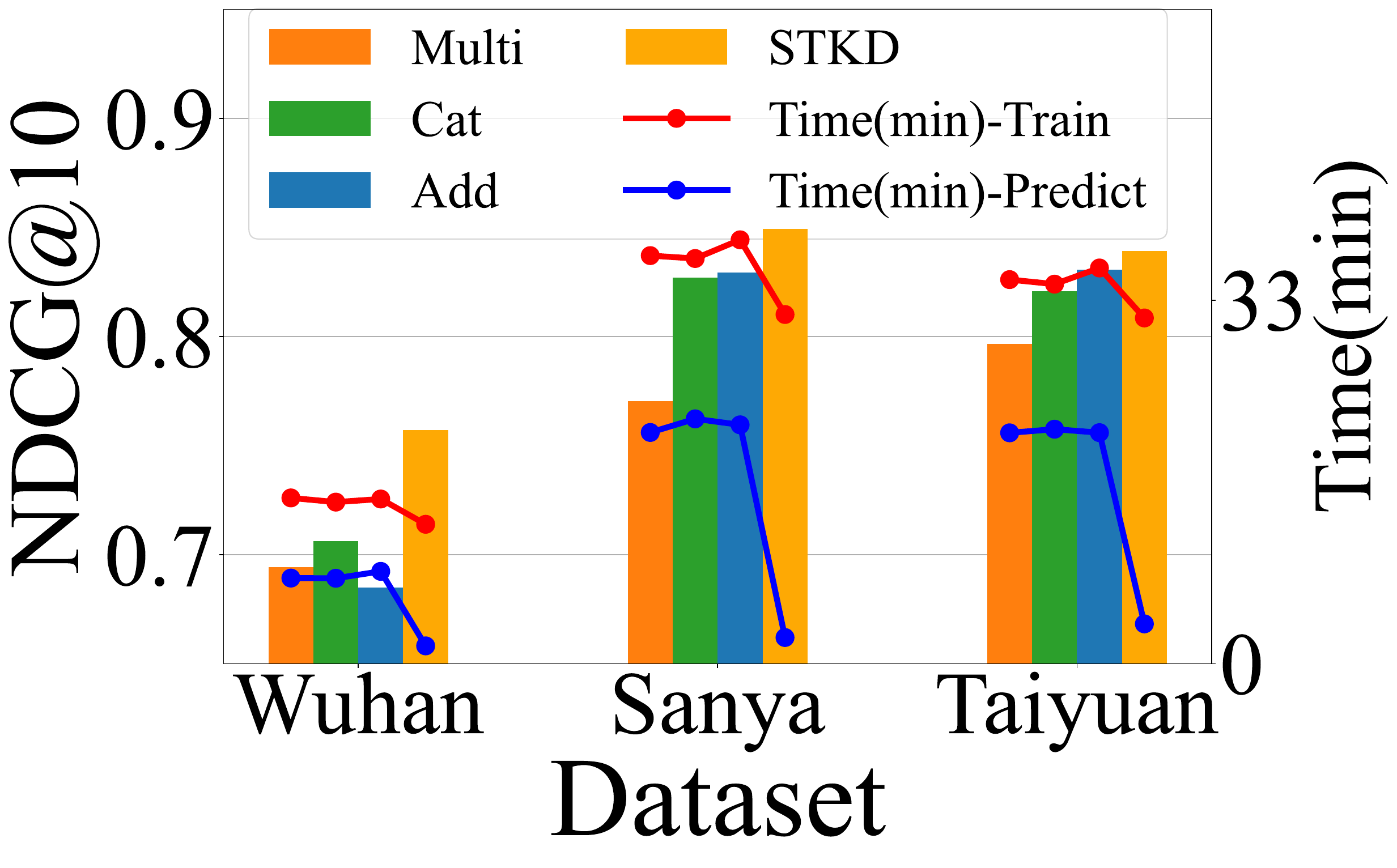}
    \end{minipage}
}
\subfigure
{
    \begin{minipage}[t]{0.47\linewidth}
        \centering
        \includegraphics[width=1.05\linewidth]{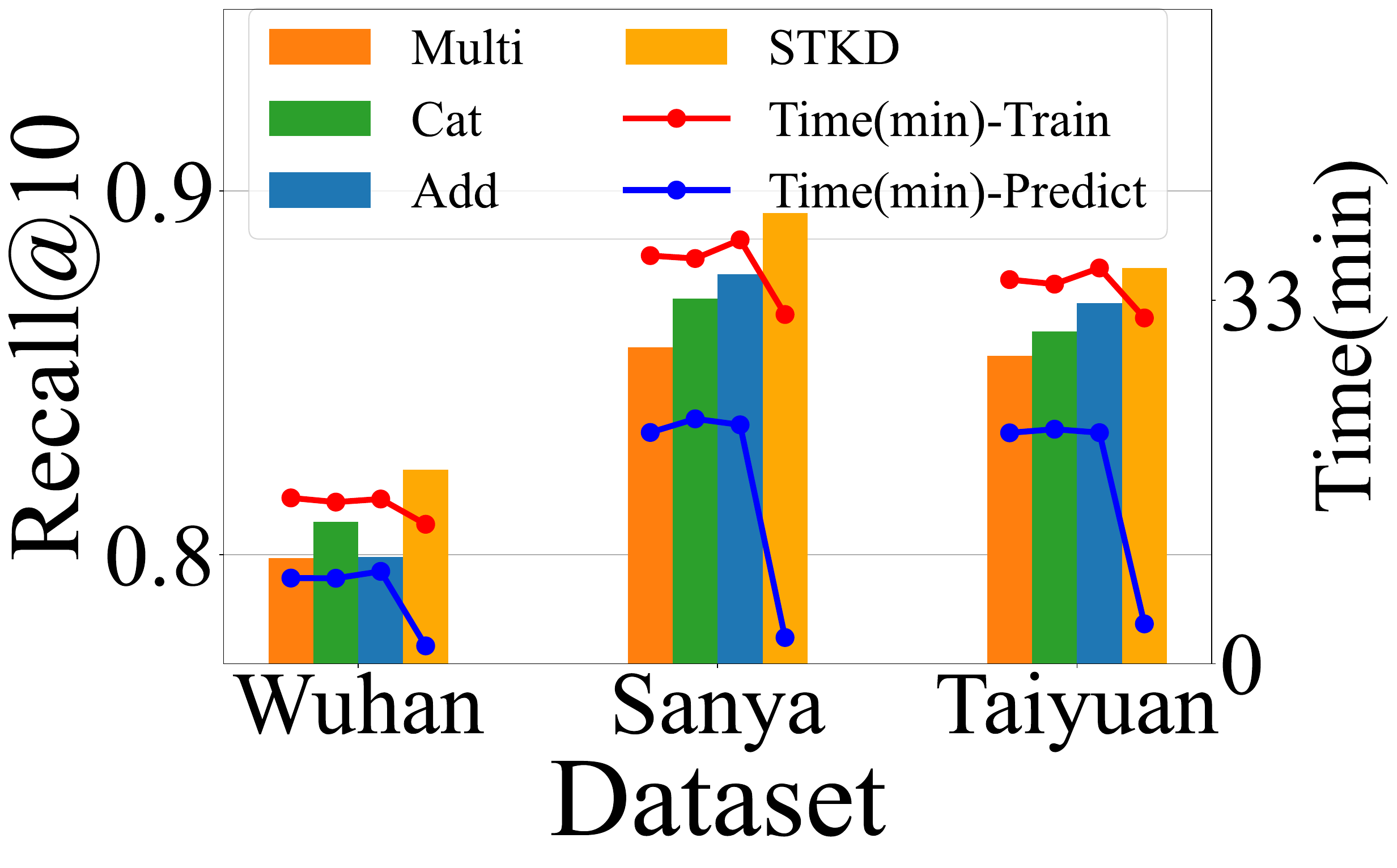}
    \end{minipage}
}
\caption{Study on different knowledge fusion methods. Multi refers to \textit{multiplication}, Cat refers to \textit{concatenation}, Add refers to \textit{addition}, STKD denotes our proposed strategy, and Time indicates model training and prediction duration.}
\label{Fig6}
\end{figure}

\subsection{Ablation Studies}
To assess the contribution of each component of STKDRec, we perform ablation studies on all datasets. The results of the STKDRec variants are shown in Table 3. \textit{-w/o SP} denotes a variant without spatial position embedding. \textit{-w/o F} and \textit{-w/o C} denote two variants considering only spatial region and spatial distance information separately. \textit{-w/o KD} denotes a variant using only the ST-Transformer module; \textit{-w/o SP + KD} denotes a variant using only the original Transformer to encode sequence. 
The results in Table 3 show that the variants \textit{-w/o SP}, \textit{-w/o F}, and \textit{-w/o C} consistently perform worse than STKDRec on all datasets, highlighting the importance of geospatial information for modeling dynamic user preferences in takeaway recommendations. The results for \textit{-w/o KD} further indicate that the STKD strategy is crucial for facilitating the fusion of spatial-temporal knowledge from STKG and sequence data. The significant drop in performance for \textit{-w/o KD+SP} demonstrates the effectiveness of all the proposed improvements.

To validate the efficacy of the STKD strategy, we replace it with other knowledge fusion strategies, including \textit{concatenation}, \textit{addition}, and \textit{multiplication}. The results are shown in Figure 3. STKDRec achieves superior performance and better training and prediction efficiency than other variants. 
This is because the knowledge distributions of the STKG encoder and the ST-Transformer differ significantly and exist in distinct vector spaces. 
Simple combining them fails to effectively integrate the differing distributions and may transform the distribution of the STKG encoder into noise, thereby damaging the distribution of the ST-Transformer.
In addition, these variants necessitate additional computations to align the distinct vector spaces, which further compromises the overall efficiency of the model.
STKD directly aligns the predicted distributions of the ST-Transformer and STKG encoder, facilitating smoother and more effective knowledge transfer. 
% STKD optimizes the student's predicted distribution to align with the teacher model, effectively integrating heterogeneous data. 
% Additionally, STKD achieves superior training efficiency by directly aligning the predicted distributions of the ST-Transformer and STKG encoder, facilitating smoother and more effective knowledge transfer. 
% In contrast, other methods involve additional fusion operations that fail to reconcile discrepancies in knowledge distributions, resulting in extended training durations and suboptimal performance.

\begin{figure}[htbp]
\centering
\subfigure{
    \begin{minipage}[t]{0.47\linewidth}
        \centering
        \includegraphics[width=1\linewidth]{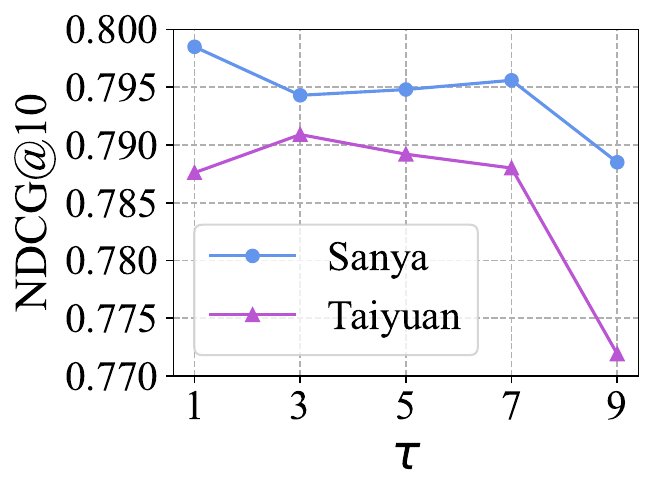}
    \end{minipage}    }
\subfigure
{
    \begin{minipage}[t]{0.47\linewidth}
        \centering
        \includegraphics[width=1\linewidth]{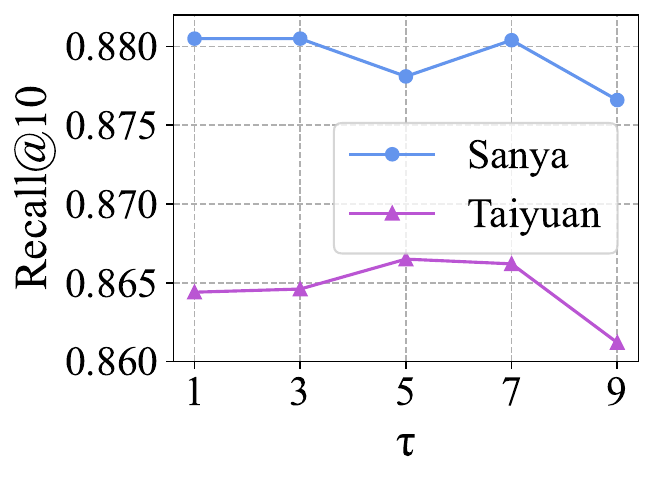}
    \end{minipage}
}\vspace{-0.4cm}
\subfigure{
    \begin{minipage}[t]{0.47\linewidth}
        \centering
        \includegraphics[width=1\linewidth]{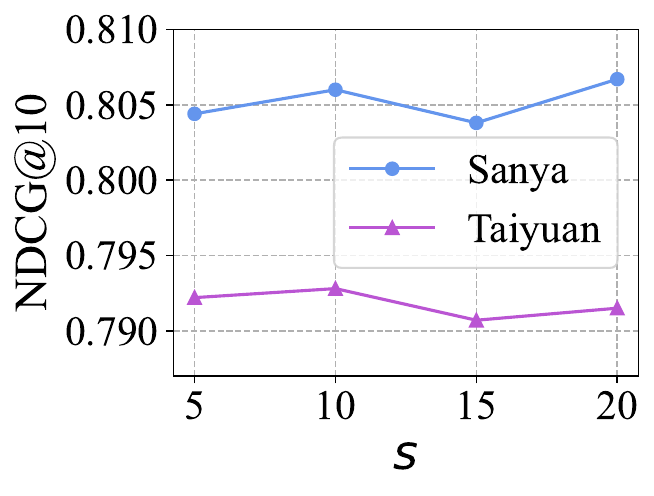}
    \end{minipage}
}
\subfigure{
    \begin{minipage}[t]{0.47\linewidth}
        \centering
        \includegraphics[width=1\linewidth]{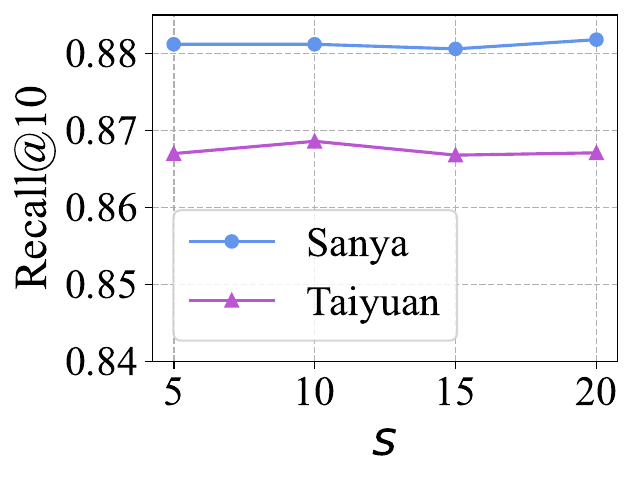}
    \end{minipage}
}
\caption{Study on different temperature $\tau$ and the number of neighbor nodes sampled $s$.}
\label{Fig6}
\end{figure}

\subsection{Parameter Sensitivity Study}
We conduct experiments on the Sanya and Taiyuan datasets to investigate the impact of two hyper-parameters: the number of neighbor nodes sampled $s$ in the STKG and the temperature coefficient $\tau$ for STKD. Figure 4 illustrates the performance of STKDRec on the Sanya and Taiyuan datasets under various settings for $s$ and $\tau$. 
For the neighbor sample size $s$, changing its value results in minimal changes in model performance, indicating that STKDRec is not sensitive to the sampling size $s$. Therefore, selecting the appropriate parameter is essential for optimal results. The optimal value for the Sanya dataset is 20, while for the Taiyuan dataset is 10.
For the temperature coefficient $\tau$, when $\tau$ exceeds 7, the model's performance significantly declines, while when $\tau$ is less than 7, the model's performance remains relatively stable.
This indicates that an excessively large coefficient weakens the teacher model's information, making it difficult for the student model to learn. 
% Therefore, selecting a moderate value for $\tau$ ensures that the teacher model's output retains sufficient information for the student model to learn effectively.
% The temperature coefficient $\tau$ generally decreases for the Sanya dataset, as a high coefficient weakens the teacher model's information, making it difficult for the student model to learn. For the Taiyuan dataset, $\tau$ first increases and then decreases, indicating that a moderate coefficient ensures the teacher model's output retains sufficient information for the student model to learn effectively.

\subsection{Case Study}
To evaluate the impact of the STKD strategy, we conduct a case study comparing the recommendation results of STKDRec and its variant $\textit{-w/o KD}$. 
Given user u420's historical purchase sequence, Figure 5 illustrates the Top-5 takeaway recommendation rankings and their corresponding probabilities for u420 at 4:40 PM on Wednesday at their workplace. 
Coffee and donuts exhibit strong spatial-temporal dependencies and frequent co-purchasing behaviors, driven by their recurrent joint consumption in workplace areas during afternoon tea time. 
The variant $\textit{-w/o KD}$ only relies on the user’s historical purchasing data, failing to account for these collaborative associations, which leads to incorrect recommendations.
However, by incorporating STKG, STKDRec can capture the spatial-temporal and collaborative relationships between coffee and donuts, allowing it to recommend the donuts for u420 accurately. This case further demonstrates the effectiveness of utilizing the STKD strategy to integrate the spatial-temporal knowledge from both STKG and sequence data.
% The variant $\textit{-w/o KD}$ only recommends pizza, which u420 has previously purchased at the workplace.
% STKDRec correctly recommends a donut because coffee and donuts have strong spatial-temporal in the STKG, as they are frequently purchased together by users in the workplace during afternoon tea time. This enables STKDRec to capture the spatial-temporal and collaborative associations between users and takeaways. As a result, it accurately recommends the donut for u420. This case further demonstrates the effectiveness of STKDRec.

\begin{figure}[h]
  \centering
  \includegraphics[width=\linewidth]{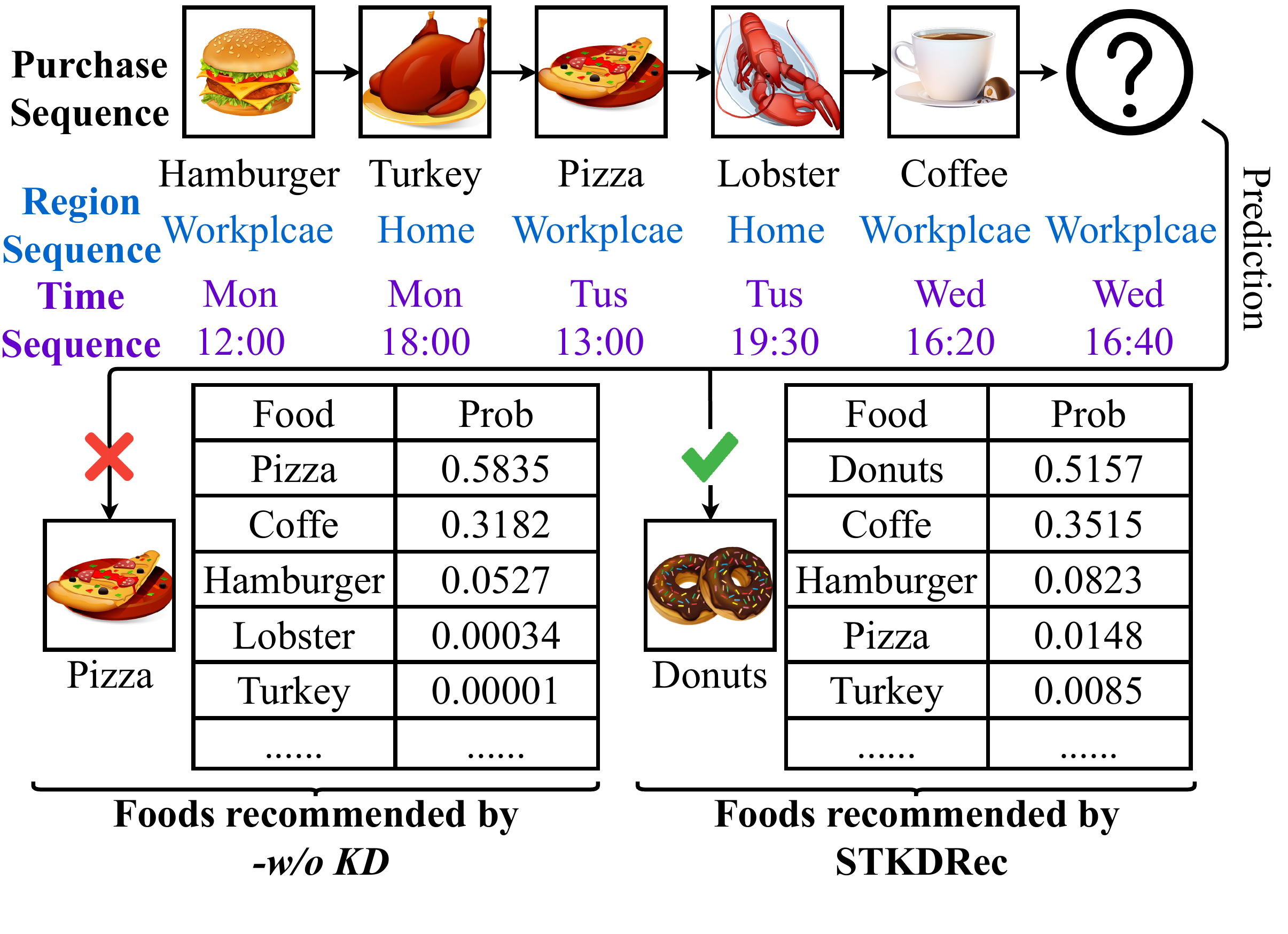}
  \caption{Visualization of case study on recommendation results.}
\end{figure}

\section{Conclusion}
In this paper, we propose a novel spatial-temporal knowledge distillation model for takeaway recommendation, termed STKDRec.
The model involves two stages: pre-training and STKD. During the pre-training stage, we train an STKG encoder to extract rich spatial-temporal knowledge from STKG. During the STKD stage, we apply an ST-Transformer to capture dynamic user preferences on fine-grained spatial region and spatial distance information from a sequential perspective. The STKD strategy is utilized to integrate heterogeneous spatial-temporal knowledge from both STKG and sequence data while reducing computational overhead.
Experimental results on three real datasets show that the performance of STKDRec is significantly better than the state-of-the-art baselines methods.

% \appendix
% \bigskip
% \noindent 
\section*{Acknowledgements}
This work was supported by the National Natural Science Foundation of China (No. 62272033).

\bibliography{aaai25}

\begin{thebibliography}{27}
\providecommand{\natexlab}[1]{#1}

\bibitem[{Chen et~al.(2022)Chen, Wan, Guo, Huang, Zheng, Li, Lin, and Lin}]{chen2022building}
Chen, W.; Wan, H.; Guo, S.; Huang, H.; Zheng, S.; Li, J.; Lin, S.; and Lin, Y. 2022.
\newblock Building and exploiting spatial--temporal knowledge graph for next POI recommendation.
\newblock \emph{Knowledge-Based Systems}, 258: 109951.

\bibitem[{Chen et~al.(2024)Chen, Wan, Wu, Zhao, Cheng, Li, and Lin}]{chen2024local}
Chen, W.; Wan, H.; Wu, Y.; Zhao, S.; Cheng, J.; Li, Y.; and Lin, Y. 2024.
\newblock Local-global history-aware contrastive learning for temporal knowledge graph reasoning.
\newblock In \emph{2024 IEEE 40th International Conference on Data Engineering (ICDE)}, 733--746. IEEE.

\bibitem[{Devlin et~al.(2018)Devlin, Chang, Lee, and Toutanova}]{devlin2018bert}
Devlin, J.; Chang, M.-W.; Lee, K.; and Toutanova, K. 2018.
\newblock Bert: Pre-training of deep bidirectional transformers for language understanding.
\newblock \emph{arXiv preprint arXiv:1810.04805}.

\bibitem[{Du et~al.(2023{\natexlab{a}})Du, Lin, Gao, Ji, Wang, Zhou, He, Jia, and Hu}]{BASM}
Du, B.; Lin, S.; Gao, J.; Ji, X.; Wang, M.; Zhou, T.; He, H.; Jia, J.; and Hu, N. 2023{\natexlab{a}}.
\newblock BASM: A bottom-up adaptive spatiotemporal model for online food ordering service.
\newblock In \emph{2023 IEEE 39th International Conference on Data Engineering (ICDE)}, 3549--3562. IEEE.

\bibitem[{Du et~al.(2023{\natexlab{b}})Du, Yuan, Zhao, Qu, Zhuang, Liu, Liu, and Sheng}]{FEARec}
Du, X.; Yuan, H.; Zhao, P.; Qu, J.; Zhuang, F.; Liu, G.; Liu, Y.; and Sheng, V.~S. 2023{\natexlab{b}}.
\newblock Frequency enhanced hybrid attention network for sequential recommendation.
\newblock In \emph{Proceedings of the 46th International ACM SIGIR Conference on Research and Development in Information Retrieval}, 78--88.

\bibitem[{Gao et~al.(2023)Gao, He, Kan, Han, Qiao, and Li}]{gao2023learning}
Gao, Y.; He, Y.; Kan, Z.; Han, Y.; Qiao, L.; and Li, D. 2023.
\newblock Learning joint structural and temporal contextualized knowledge embeddings for temporal knowledge graph completion.
\newblock In \emph{Findings of the Association for Computational Linguistics: ACL 2023}, 417--430.

\bibitem[{Hamilton, Ying, and Leskovec(2017)}]{STKGsample}
Hamilton, W.; Ying, Z.; and Leskovec, J. 2017.
\newblock Inductive representation learning on large graphs.
\newblock \emph{Advances in neural information processing systems}, 30.

\bibitem[{Hidasi et~al.(2015)Hidasi, Karatzoglou, Baltrunas, and Tikk}]{GRU4Rec}
Hidasi, B.; Karatzoglou, A.; Baltrunas, L.; and Tikk, D. 2015.
\newblock Session-based recommendations with recurrent neural networks.
\newblock \emph{arXiv preprint arXiv:1511.06939}.

\bibitem[{Hinton, Vinyals, and Dean(2015)}]{hinton2015distilling}
Hinton, G.; Vinyals, O.; and Dean, J. 2015.
\newblock Distilling the knowledge in a neural network.
\newblock \emph{arXiv preprint arXiv:1503.02531}.

\bibitem[{Kang et~al.(2021)Kang, Hwang, Kweon, and Yu}]{kang2021topology}
Kang, S.; Hwang, J.; Kweon, W.; and Yu, H. 2021.
\newblock Topology distillation for recommender system.
\newblock In \emph{Proceedings of the 27th ACM SIGKDD Conference on Knowledge Discovery \& Data Mining}, 829--839.

\bibitem[{Kang et~al.(2023)Kang, Kweon, Lee, Lian, Xie, and Yu}]{kang2023distillation}
Kang, S.; Kweon, W.; Lee, D.; Lian, J.; Xie, X.; and Yu, H. 2023.
\newblock Distillation from heterogeneous models for top-k recommendation.
\newblock In \emph{Proceedings of the ACM Web Conference 2023}, 801--811.

\bibitem[{Kang and McAuley(2018)}]{SASRec}
Kang, W.-C.; and McAuley, J. 2018.
\newblock Self-attentive sequential recommendation.
\newblock In \emph{2018 IEEE international conference on data mining (ICDM)}, 197--206. IEEE.

\bibitem[{Lin et~al.(2022)Lin, Yu, Ji, Zhou, He, Sang, Jia, Cao, and Hu}]{StEN}
Lin, S.; Yu, Y.; Ji, X.; Zhou, T.; He, H.; Sang, Z.; Jia, J.; Cao, G.; and Hu, N. 2022.
\newblock Spatiotemporal-enhanced network for click-through rate prediction in location-based services.
\newblock \emph{arXiv preprint arXiv:2209.09427}.

\bibitem[{Liu, Zhu, and Wu(2023)}]{KGDPL}
Liu, H.; Zhu, Y.; and Wu, Z. 2023.
\newblock Knowledge graph-based behavior denoising and preference learning for sequential recommendation.
\newblock \emph{IEEE Transactions on Knowledge and Data Engineering}.

\bibitem[{Qiu et~al.(2022)Qiu, Huang, Yin, and Wang}]{DuoRec}
Qiu, R.; Huang, Z.; Yin, H.; and Wang, Z. 2022.
\newblock Contrastive learning for representation degeneration problem in sequential recommendation.
\newblock In \emph{Proceedings of the fifteenth ACM international conference on web search and data mining}, 813--823.

\bibitem[{Rendle, Freudenthaler, and Schmidt-Thieme(2010)}]{Markov_model}
Rendle, S.; Freudenthaler, C.; and Schmidt-Thieme, L. 2010.
\newblock Factorizing personalized markov chains for next-basket recommendation.
\newblock In \emph{Proceedings of the 19th international conference on World wide web}, 811--820.

\bibitem[{Shi et~al.(2024)Shi, Yang, Lv, Yuan, Kou, Luo, and Xu}]{shi2024self}
Shi, L.; Yang, J.; Lv, P.; Yuan, L.; Kou, F.; Luo, J.; and Xu, M. 2024.
\newblock Self-derived knowledge graph contrastive learning for recommendation.
\newblock In \emph{Proceedings of the 32nd ACM International Conference on Multimedia}, 7571--7580.

\bibitem[{Shin et~al.(2024)Shin, Choi, Wi, and Park}]{BSARec}
Shin, Y.; Choi, J.; Wi, H.; and Park, N. 2024.
\newblock An attentive inductive bias for sequential recommendation beyond the self-attention.
\newblock In \emph{Proceedings of the AAAI Conference on Artificial Intelligence}, volume~38, 8984--8992.

\bibitem[{Sun et~al.(2019)Sun, Liu, Wu, Pei, Lin, Ou, and Jiang}]{Bert4Rec}
Sun, F.; Liu, J.; Wu, J.; Pei, C.; Lin, X.; Ou, W.; and Jiang, P. 2019.
\newblock BERT4Rec: Sequential recommendation with bidirectional encoder representations from transformer.
\newblock In \emph{Proceedings of the 28th ACM international conference on information and knowledge management}, 1441--1450.

\bibitem[{Tang and Wang(2018)}]{Caser}
Tang, J.; and Wang, K. 2018.
\newblock Personalized top-n sequential recommendation via convolutional sequence embedding.
\newblock In \emph{Proceedings of the eleventh ACM international conference on web search and data mining}, 565--573.

\bibitem[{Wu et~al.(2023)Wu, Wan, Chen, Wu, Shen, and Lin}]{wu2023towards}
Wu, S.; Wan, H.; Chen, W.; Wu, Y.; Shen, J.; and Lin, Y. 2023.
\newblock Towards enhancing relational rules for knowledge graph link prediction.
\newblock \emph{arXiv preprint arXiv:2310.13411}.

\bibitem[{Xia et~al.(2022)Xia, Yin, Yu, Wang, Xu, and Nguyen}]{xia2022device}
Xia, X.; Yin, H.; Yu, J.; Wang, Q.; Xu, G.; and Nguyen, Q. V.~H. 2022.
\newblock On-device next-item recommendation with self-supervised knowledge distillation.
\newblock In \emph{Proceedings of the 45th International ACM SIGIR Conference on Research and Development in Information Retrieval}, 546--555.

\bibitem[{Ye, Xia, and Huang(2023)}]{MAERec}
Ye, Y.; Xia, L.; and Huang, C. 2023.
\newblock Graph masked autoencoder for sequential recommendation.
\newblock In \emph{Proceedings of the 46th International ACM SIGIR Conference on Research and Development in Information Retrieval}, 321--330.

\bibitem[{Zhang et~al.(2021)Zhang, Liu, Sun, and Shah}]{GLNN}
Zhang, S.; Liu, Y.; Sun, Y.; and Shah, N. 2021.
\newblock Graph-less neural networks: Teaching old mlps new tricks via distillation.
\newblock \emph{arXiv preprint arXiv:2110.08727}.

\bibitem[{Zhang et~al.(2022)Zhang, Liu, Xu, Xiong, Lei, He, Cui, and Miao}]{GCL4SR}
Zhang, Y.; Liu, Y.; Xu, Y.; Xiong, H.; Lei, C.; He, W.; Cui, L.; and Miao, C. 2022.
\newblock Enhancing sequential recommendation with graph contrastive learning.
\newblock \emph{arXiv preprint arXiv:2205.14837}.

\bibitem[{Zhang et~al.(2023)Zhang, Wu, Le, Zhu, Zhuang, Han, Li, Lin, An, and Xu}]{meituan}
Zhang, Y.; Wu, Y.; Le, R.; Zhu, Y.; Zhuang, F.; Han, R.; Li, X.; Lin, W.; An, Z.; and Xu, Y. 2023.
\newblock Modeling dual period-varying preferences for takeaway recommendation.
\newblock In \emph{Proceedings of the 29th ACM SIGKDD Conference on Knowledge Discovery and Data Mining}, 5628--5638.

\bibitem[{Zhu et~al.(2021)Zhu, Chen, Wu, Liu, and Zhao}]{zhu2021combining}
Zhu, Q.; Chen, X.; Wu, P.; Liu, J.; and Zhao, D. 2021.
\newblock Combining curriculum learning and knowledge distillation for dialogue generation.
\newblock In \emph{Findings of the Association for Computational Linguistics: EMNLP 2021}, 1284--1295.

\end{thebibliography}

\end{document}